\documentclass[runningheads]{llncs}
\usepackage{graphicx}
\usepackage{amsmath,amssymb} 
\usepackage{color}
\usepackage{caption}
\usepackage{adjustbox}
\usepackage{multirow}
\usepackage{subfigure}
\usepackage{color}

\usepackage[width=122mm,left=12mm,paperwidth=146mm,height=193mm,top=12mm,paperheight=217mm]{geometry}
\begin{document}
\pagestyle{headings}
\mainmatter

\title{Detecting Text in Natural Image with Connectionist Text Proposal  Network} 

\titlerunning{Detecting Text in Natural Image with CTPN}

\authorrunning{Z. Tian, W. Huang, T. He, P. He and Y. Qiao}

\author{Zhi Tian$^1$, Weilin Huang\thanks{Corresponding author}$^{1,2}$, Tong He$^1$, Pan He$^1$,  and Yu Qiao$^{1,3}$}


\institute{$^{1}${Shenzhen Key Lab of Comp. Vis and Pat. Rec., \\
Shenzhen Institutes of Advanced Technology, Chinese Academy of Sciences}\\
$^{2}$University of Oxford {} {}
$^{3}$The Chinese University of Hong Kong\\
\email{ \{zhi.tian;wl.huang;tong.he;pan.he;yu.qiao}@siat.ac.cn\}
}

\maketitle

\begin{abstract}
We propose a novel Connectionist Text Proposal Network (CTPN) that  \textit{accurately} localizes text lines in natural image.  The CTPN detects a text line in a sequence of  fine-scale text proposals directly in convolutional feature maps.  We develop a vertical anchor mechanism that jointly predicts  location and text/non-text score  of each fixed-width proposal,   considerably improving localization accuracy. The sequential proposals  are naturally connected by a recurrent neural network, which is seamlessly incorporated into the convolutional network, resulting in an end-to-end trainable model. This allows the CTPN to explore rich context information of image, making it powerful to detect extremely ambiguous text.
The CTPN works reliably on multi-scale and multi-language text without further post-processing, departing from previous bottom-up methods requiring multi-step post filtering. It achieves 0.88 and 0.61 F-measure on the ICDAR 2013 and 2015 benchmarks, surpassing recent results \cite{Gupta2016,Zhang2016} by a large margin. The CTPN is computationally efficient with $0.14s/$image, by using the very deep VGG16 model \cite{Simonyan2015}. 
Online demo is available at: \textcolor{blue}{http://textdet.com/}. 

\end{abstract}

\section{Introduction}

Reading text in natural image has recently attracted increasing attention in computer vision \cite{Gupta2016,Huang2014,Jaderberg2015,He2017,Zhang2016,He2016,He2015,Busta2015,Tian2015,Yin2015}. This is due to its numerous practical applications such as image OCR, multi-language translation,  image retrieval, etc. It includes two sub tasks: text detection and recognition. This work focus on the detection task \cite{Huang2014,Busta2015,Tian2015,Yin2015}, which is more challenging than recognition task carried out on a well-cropped word image \cite{Jaderberg2015,He2015}. Large variance of text patterns and highly cluttered background pose main challenge of \textit{accurate} text localization.

Current approaches for text detection mostly employ a bottom-up pipeline \cite{Tian2015,Busta2015,Huang2014,Yin2015,Yin2014}. They commonly start from  low-level character or stroke detection, which is typically followed by a number of subsequent steps: non-text component filtering, text line construction and text line verification.  These multi-step bottom-up approaches are generally complicated with less  robustness and reliability. Their performance heavily rely on the results of character detection, and connected-components methods or sliding-window methods have been proposed. These methods commonly explore low-level features (e.g., based on SWT \cite{Epshtein2010,Huang2013}, MSER \cite{Huang2014,Yin2014,Neumann2015}, or HoG \cite{Tian2015}) to distinguish text candidates from background. However, they are not robust by identifying individual strokes or characters separately, without context information. For example, it is more confident for people to identify a sequence of characters than an individual one, especially when a character is extremely ambiguous. These limitations often result in a large number of non-text components in character detection, causing  main difficulties for handling them in following steps. Furthermore, these false detections are easily accumulated sequentially in  bottom-up pipeline, as pointed out in \cite{Tian2015}. To address these problems, we exploit strong deep features for detecting text information  directly in  convolutional maps. We develop text anchor mechanism that \textit{accurately} predicts text locations in fine scale. Then, an in-network recurrent architecture is proposed to connect these fine-scale text proposals in sequences, allowing them to encode rich context information.

 \begin{figure}[tb]
\centering
\subfigure[]{\includegraphics[height=3cm,width=8.6cm]{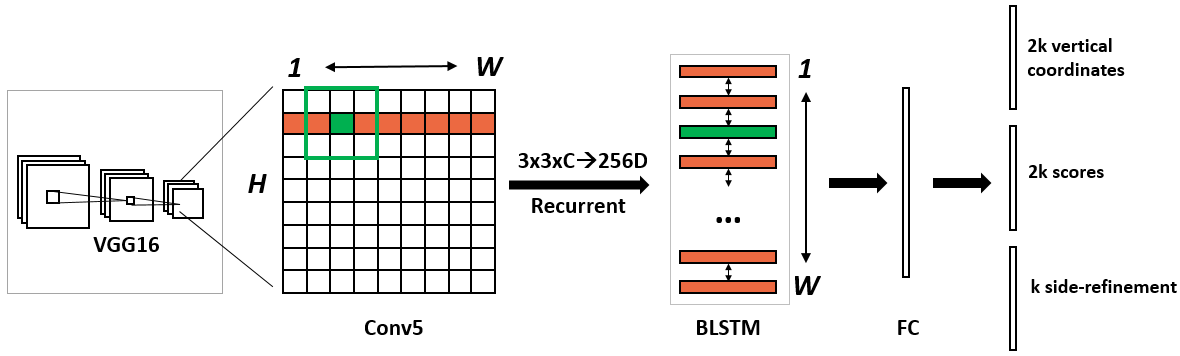}}
\subfigure[]{\includegraphics[height=2.7cm,width=3.2cm]{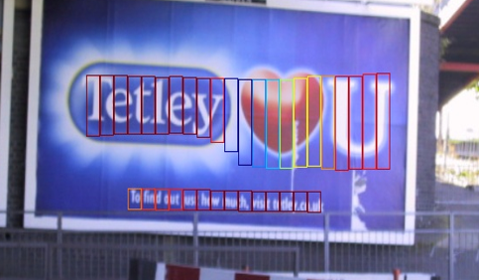}}
\caption{(a) Architecture of the Connectionist Text Proposal Network (CTPN). We densely slide a 3$\times$3 spatial window through the last convolutional maps (\textit{conv5}) of the VGG16 model \cite{Simonyan2015}. The sequential windows in each row are recurrently connected by a Bi-directional LSTM (BLSTM) \cite{Graves2005}, where the  convolutional feature (3$\times$3$\times$C) of each window is used as input of the 256D BLSTM (including two 128D LSTMs). The RNN layer is connected to a 512D fully-connected layer, followed by the output layer, which jointly predicts text/non-text scores, $y$-axis coordinates  and side-refinement offsets of $k$ anchors. (b) The CTPN outputs sequential fixed-width fine-scale text proposals.  Color of  each box indicates the text/non-text score. Only the boxes with positive scores are presented.}
\label{fig:main}
\end{figure}



Deep Convolutional Neural Networks (CNN) have recently advanced general object detection substantially \cite{Ren2015,Girshick2015,Girshick2014}. 
The state-of-the-art method is Faster Region-CNN (R-CNN) system \cite{Ren2015} where a Region Proposal Network (RPN) is proposed to  generate  high-quality class-agnostic object proposals directly from convolutional feature maps. Then the RPN proposals are fed into a Fast R-CNN \cite{Girshick2015} model for further classification and refinement, leading to the  state-of-the-art performance on generic object detection.
\textit{However, it is difficult to apply these general object detection systems directly to scene text detection, which generally requires a higher localization accuracy}.  
In generic object detection, each object has a well-defined closed boundary \cite{Cheng2014}, while such a well-defined boundary may not exist in text, since a text line or word is composed of a number of separate characters or strokes. For object detection, a typical correct detection is defined loosely, e.g., by an  overlap  of $>0.5$ between the detected bounding box and its ground truth (e.g., the PASCAL standard \cite{Everingham2010}), since people can recognize an object easily from major part of it.
By contrast, reading text comprehensively is a fine-grained recognition task which requires a correct detection that covers a full region of a text line or word. Therefore, text detection generally requires a more \textit{accurate} localization, leading to a different evaluation standard, e.g., the Wolf's standard \cite{Wolf2006} which is commonly employed by text  benchmarks \cite{Karatzas2013,Minetto2010}.

In this work, we fill this gap by extending the RPN architecture \cite{Ren2015} to \textit{accurate} text line localization.  We present several technical developments that tailor generic object detection model elegantly towards our problem. We strive for a further step by proposing an in-network recurrent mechanism that allows our model to detect text sequence  directly in the convolutional maps, avoiding further post-processing by an additional costly CNN detection model.




\subsection{Contributions}
We  propose  a novel Connectionist Text Proposal Network (CTPN) that directly localizes text sequences in  convolutional layers.
This overcomes a number of main limitations raised by previous bottom-up approaches building on character detection. We leverage the advantages of  strong deep convolutional features and sharing computation mechanism, and propose the CTPN architecture which is described in Fig. \ref{fig:main}. It makes the following major contributions:


First, we cast the problem of text detection into localizing a sequence of fine-scale text proposals. We develop an anchor  regression mechanism that jointly predicts  vertical location and text/non-text score of each text proposal, resulting in  an excellent localization accuracy. This departs from the RPN prediction of a whole object, which is difficult to provide a satisfied localization accuracy.

Second, we propose an in-network recurrence mechanism that elegantly connects sequential text proposals  in the convolutional feature maps. This connection allows our detector to explore meaningful context information of text line, making it powerful to detect  extremely challenging text reliably.


Third, both methods are integrated seamlessly to meet the nature of text sequence, resulting in a unified end-to-end trainable model. Our method is able to handle multi-scale and multi-lingual text in a single process, avoiding further post filtering or refinement. 

Fourth, our method achieves new state-of-the-art results on a number of benchmarks, significantly improving recent results (e.g., 0.88 F-measure over 0.83 in \cite{Gupta2016} on the ICDAR 2013, and 0.61 F-measure over 0.54 in \cite {Zhang2016} on the ICDAR 2015). Furthermore,  it is computationally efficient, resulting in a $0.14s/$image running time (on the ICDAR 2013) by using the very deep VGG16 model \cite{Simonyan2015}.

\section{Related Work}
\textbf{Text detection.}  Past works in scene text detection have been dominated by bottom-up approaches which are generally built on stroke or character detection. They can be roughly grouped into two categories, connected-components (CCs) based approaches and sliding-window based methods.  The CCs based approaches discriminate text and non-text pixels by using a fast filter, and then text pixels are greedily grouped into stroke or character candidates, by using  low-level properties, e.g., intensity, color, gradient, etc. \cite{Yin2014,Huang2014,Yin2015,Huang2013,Epshtein2010}.  The sliding-window based methods detect character candidates by densely moving a multi-scale window through an image. The character or non-character window is discriminated by a pre-trained classifier, by using manually-designed features \cite{Tian2015,Wang2011},  or recent CNN features \cite{Jaderberg2014}.  However, both groups of methods commonly suffer from poor performance of character detection, causing accumulated errors in following component filtering and text line construction steps. Furthermore, robustly filtering out non-character components or confidently verifying detected text lines are even difficult themselves \cite {Busta2015,Yin2014,Huang2014}. Another limitation is that the sliding-window methods  are computationally expensive, by running a classifier on a huge number of the sliding windows. 


\textbf{Object detection.}  Convolutional Neural Networks (CNN) have recently advanced general object detection substantially \cite{Ren2015,Girshick2015,Girshick2014}. A common strategy is to generate a number of  object proposals by employing inexpensive  low-level features, and then a strong CNN classifier is applied to further classify and refine the generated proposals. Selective Search (SS) \cite{Everingham2010} which  generates class-agnostic object proposals,  is one of the most popular methods applied in recent leading object detection systems, such as Region CNN (R-CNN) \cite{Girshick2014} and its extensions \cite{Girshick2015}. Recently, Ren \textit{et al.} \cite{Ren2015} proposed a Faster R-CNN system for object detection. They proposed a Region Proposal Network (RPN) that generates  high-quality class-agnostic object proposals  directly from the convolutional feature maps. The RPN is fast by sharing convolutional computation. However, the RPN proposals are not discriminative, and require a further refinement and classification by an additional costly CNN model, e.g., the Fast R-CNN model \cite{Girshick2015}. More importantly,  text is different significantly from general objects, making it difficult to directly apply general object detection system to this highly domain-specific task.

\section{Connectionist Text Proposal Network}
This section presents  details of the Connectionist Text Proposal Network (CTPN). It includes three key contributions that make it reliable and accurate for text localization: detecting text in fine-scale proposals,  recurrent connectionist text proposals, and  side-refinement. 

\subsection{Detecting Text in Fine-scale Proposals}
 \begin{figure}[tb]
\centering
\includegraphics[height=2cm,width=12cm]{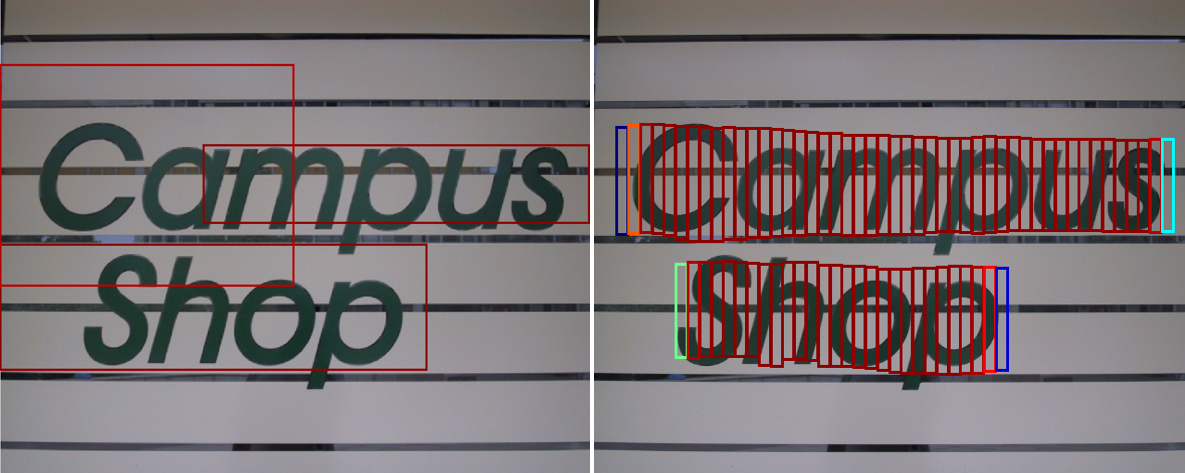}
\caption{\textbf{Left}: RPN proposals. \textbf{Right}: Fine-scale text proposals.}
\label{fig:RPN}
\end{figure}

Similar to Region Proposal Network (RPN) \cite{Ren2015}, the CTPN is essentially a fully convolutional network that allows an input image of arbitrary size. It detects a text line by densely sliding a small window in the convolutional feature maps, and outputs a sequence of fine-scale (e.g., fixed 16-pixel width) text proposals, as shown in Fig. \ref{fig:main} (b). 

We take the very deep 16-layer vggNet (VGG16) \cite{Simonyan2015} as an example to describe our approach, which is readily applicable to other deep models. Architecture of the CTPN is presented in Fig. \ref{fig:main} (a). We use a small spatial window, 3$\times$3, to slide the feature maps of last convolutional layer (e.g., the \textit{conv5} of the VGG16). The size of \textit{conv5} feature maps is determined by the size of input image, while the total stride and receptive field are fixed as 16 and 228 pixels, respectively.
Both the total stride and receptive field are fixed by the network architecture. 
Using a sliding window in the convolutional layer allows it to share convolutional computation, which is the key to reduce  computation of the costly sliding-window based methods.  


Generally,  sliding-window methods adopt multi-scale windows to detect objects of different sizes, where one window scale is fixed to objects of similar size. In \cite{Ren2015}, Ren \textit{et al.} proposed an efficient anchor  regression mechanism that allows the RPN to detect multi-scale objects with a single-scale window. The key insight is that a single window is able to predict objects in a wide range
of scales and aspect ratios, by using a number of flexible anchors.  We wish to extend this efficient anchor mechanism to our text task. 
However, text differs from generic objects substantially, which generally have a well-defined enclosed boundary and center, allowing inferring whole object from even a part of it \cite{Cheng2014}.  Text is a sequence which does not have an obvious closed boundary. It may include multi-level components, such as stroke,  character, word, text line and text region, which are not distinguished clearly between each other.  Text detection is defined in word or text line level, so that it may be easy to make an incorrect detection by defining it as  a single object, e.g., detecting part of a word.
Therefore, directly predicting the location of a text line or word may be difficult or unreliable, making it hard to get a satisfied accuracy. An example is shown in Fig. \ref{fig:RPN}, where the RPN is directly trained for localizing text lines in an image.

We look for a unique property of text that is able to generalize well to text components in all levels. 
We observed that word detection by the RPN is difficult to accurately predict the horizontal sides of words, since each character within a word is isolated or separated, making it confused to find the start and end locations of a word. 
Obviously, a text line is  a sequence which is the main difference between text and generic objects.  It is natural to consider a text line as a sequence of fine-scale text proposals, where each proposal  generally  represents a small part of a text line, e.g., a text piece with 16-pixel width. Each proposal may include a single or multiple strokes, a part of a character, a single or multiple characters, etc. 
We believe that it would be more accurate to just predict the vertical location of each proposal, by fixing its horizontal location which may be more difficult to predict.
This reduces the search space, compared to the RPN which predicts 4 coordinates of an object. 
We develop a vertical anchor mechanism that simultaneously predicts a text/non-text score and $y$-axis location of each fine-scale proposal. It is also more reliable to detect a general fixed-width text proposal than identifying  an isolate character, which is easily confused with part of a character or multiple characters.  Furthermore, detecting a text line in a sequence of fixed-width text proposals also works reliably on text of  multiple scales and multiple aspect ratios.

To this end, we design the fine-scale text proposal as follow. Our detector investigates each spatial location in the \textit{conv5} densely. A text proposal is defined to have a fixed width of 16 pixels (in the input image). This is equal to move the detector densely through the \textit{conv5} maps, where the total stride is exactly 16 pixels. Then we design $k$ vertical anchors to  predict  $y$-coordinates for each proposal. The $k$ anchors have a same horizontal location with a fixed width of 16 pixels, but their  vertical locations are varied in $k$ different heights. In our experiments, we use ten anchors for each proposal, $k=10$, whose heights are varied from 11 to 273 pixels (by $\div 0.7$ each time) in the input image. The explicit vertical coordinates are measured by the height   and $y$-axis center of a proposal bounding box. We compute relative predicted vertical coordinates ($\textbf{v}$) with respect to the bounding box location of an anchor as, 
 \begin{eqnarray}
v_c&=&(c_y-c_y^a)/h^a, \qquad v_h=\log (h/h^a) \\\label{eq:coordinate}
v^*_c&=&(c^*_y-c_y^a)/h^a, \qquad v^*_h=\log (h^*/h^a)\label{eq:coordinate_gt}
\end{eqnarray}
where $\textbf{v}=\{v_c,v_h\}$ and  $\textbf{v}^*=\{v^*_c,v^*_h\}$ are the relative predicted coordinates and ground truth coordinates, respectively. $c_y^a$ and $h^a$ are the center ($y$-axis) and height of the anchor box, which can be pre-computed from an input image. $c_y$ and $h$ are the predicted $y$-axis coordinates in the input image, while $c^*_y$ and $h^*$ are the ground truth coordinates. Therefore, each predicted text proposal has a bounding box with size of $h\times 16$ (in the input image), as shown in Fig. \ref{fig:main} (b) and Fig. \ref{fig:RPN} (right).  Generally, an text proposal is largely smaller than its effective receptive field which is 228$\times$228.

The detection processing is summarised as follow. Given an input image, we have $W \times H \times C$  \textit{conv5} features maps (by using the VGG16 model), where $C$ is the number of feature maps or channels, and $W \times H$ is the spatial arrangement. When our detector is  sliding a 3$\times$3 window densely through the conv5, each sliding-window takes a convolutional feature of $3 \times 3 \times C$ for producing the prediction. For each prediction, the horizontal location ($x$-coordinates) and $k$-anchor locations  are fixed, which can be pre-computed by mapping the spatial window location in the \textit{conv5} onto the input image. Our detector outputs the text/non-text scores and the predicted $y$-coordinates ($\textbf{v}$) for $k$ anchors at each window location.
The detected text proposals are generated from the anchors having a text/non-text score of $>0.7$  (with non-maximum suppression).
By the designed vertical anchor and fine-scale detection strategy, our detector is able to handle text lines in a wide range of scales and aspect ratios by using a single-scale image. This further reduces its computation, and at the same time, predicting accurate localizations of the text lines. Compared to the RPN or Faster R-CNN system \cite{Ren2015}, our fine-scale detection provides more detailed supervised information that naturally leads to a more accurate detection.

\subsection{Recurrent Connectionist Text Proposals} 
\begin{figure}[tb]
\centering
\includegraphics[height=2.5cm,width=3.9cm]{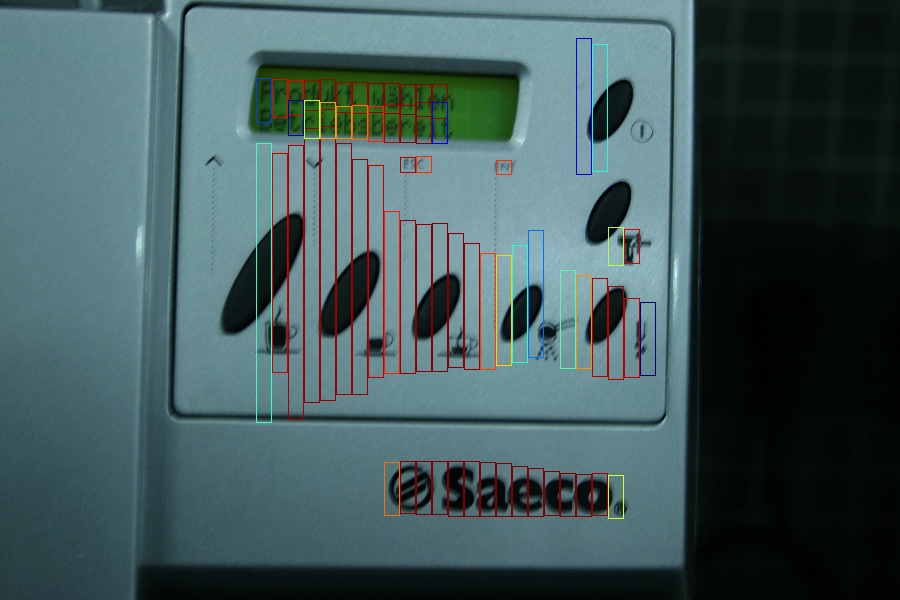}
\includegraphics[height=2.5cm,width=3.9cm]{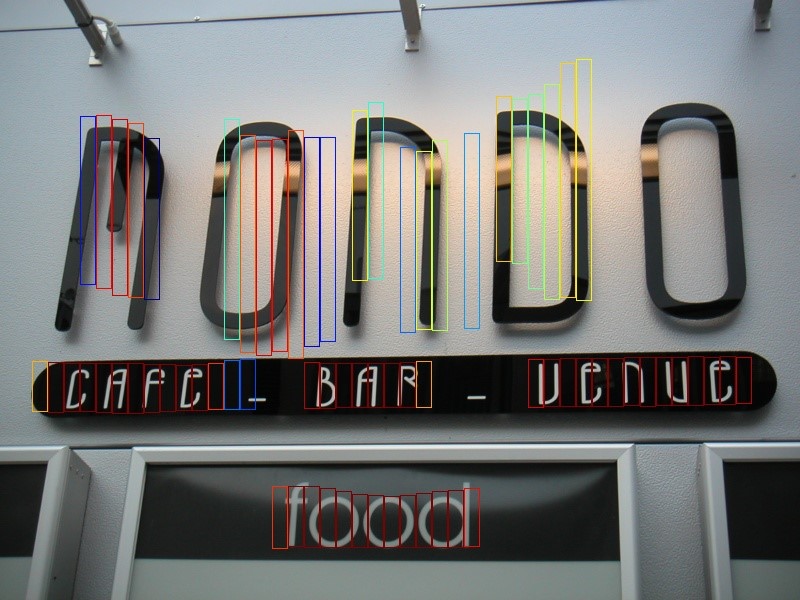}
\includegraphics[height=2.5cm,width=3.9cm]{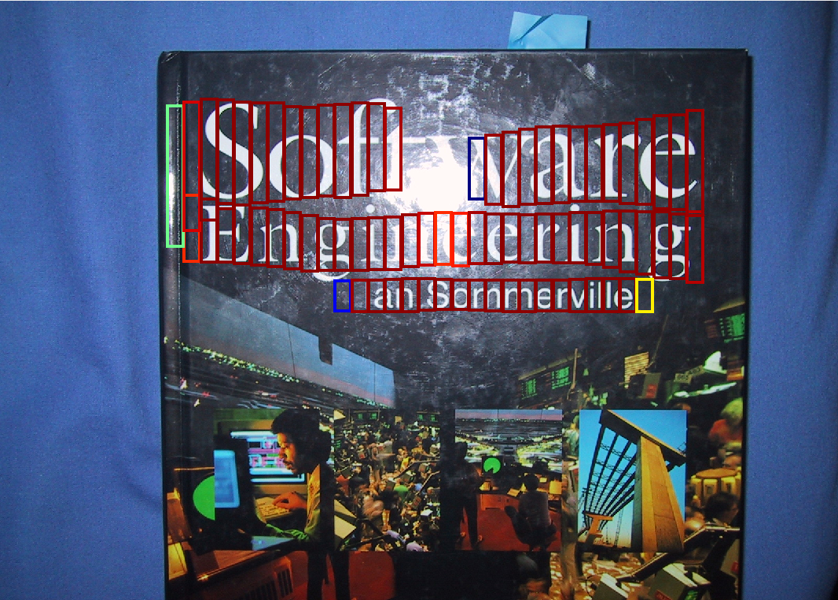}\\

\includegraphics[height=2.5cm,width=3.9cm]{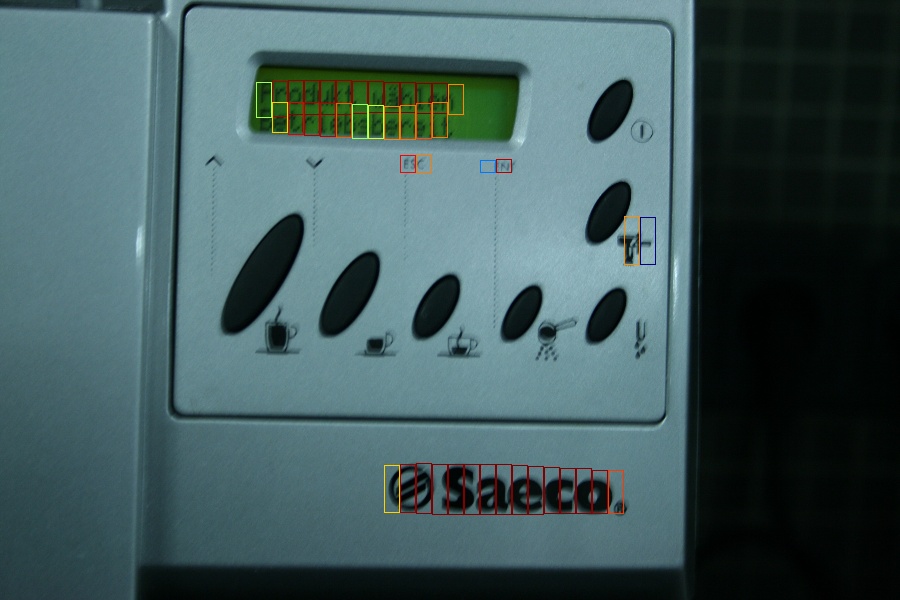}
\includegraphics[height=2.5cm,width=3.9cm]{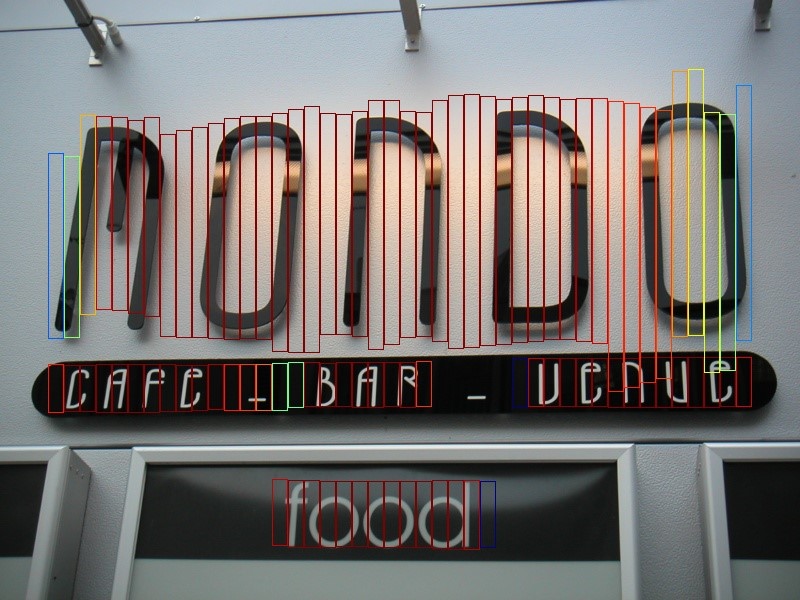}
\includegraphics[height=2.5cm,width=3.9cm]{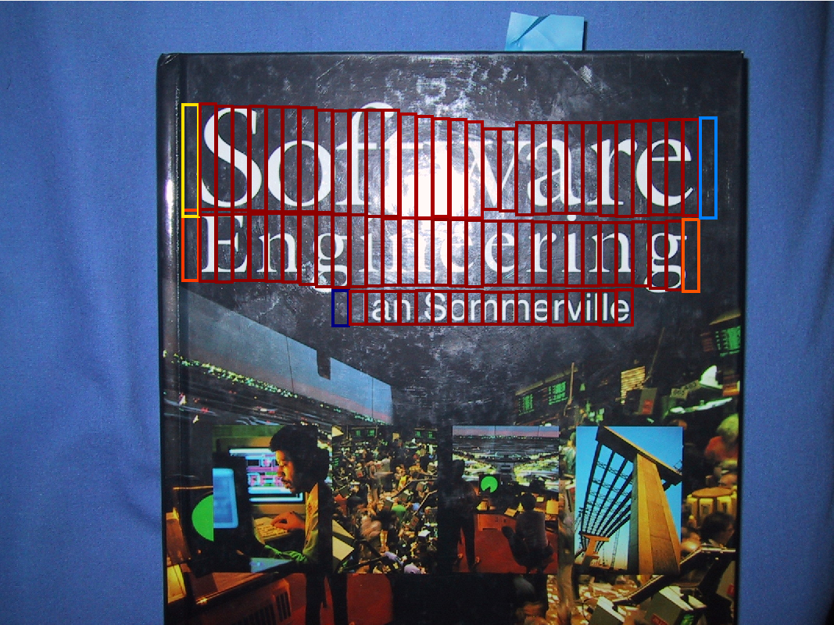}
\caption{\textbf{Top}: CTPN without RNN. \textbf{Bottom}: CTPN with RNN connection.}
\label{fig:lstm}
\end{figure}

To improve localization accuracy, we split a text line into a sequence of fine-scale text proposals, and predict each of them separately. Obviously, it is not robust to regard each isolated proposal independently. This may lead to a number of false detections on non-text objects which have a similar structure as text patterns,  such as windows, bricks, leaves, etc. (referred as text-like outliers in \cite{Huang2013}). It is also possible to discard some ambiguous  patterns which contain weak text information. Several examples are presented in Fig. \ref{fig:lstm} (top). Text have  strong sequential characteristics where the sequential context information is crucial to make a reliable decision. This has been verified by recent work \cite{He2015} where a recurrent neural network (RNN) is applied to encode this context information for text recognition. Their results have shown that the sequential context information is greatly facilitate the recognition task on cropped word images. 

Motivated from this work, we believe that this context information may also be of importance for our detection task. 
Our detector should be able to explore this important context information to make a more reliable decision, when it works on each individual proposal.
Furthermore, we aim to encode this information directly in the convolutional layer, resulting in an elegant and seamless in-network connection of the fine-scale text proposals.
RNN provides a natural choice for encoding this information recurrently using its hidden layers. To this end, we propose to design a RNN layer upon the \textit{conv5}, which takes the convolutional feature of each window  as sequential inputs, and updates its internal state recurrently  in the hidden layer, $H_t$,
\begin{eqnarray}
H_{t}=\varphi(H_{t-1}, X_t),  \qquad t=1,2,...,W
\end{eqnarray}
where $X_t \in R^{3\times 3 \times C}$ is the input \textit{conv5} feature from $t$-th  sliding-window (3$\times$3). The sliding-window moves densely from left to right, resulting in $t=1,2,...,W$ sequential features for each row.  $W$ is the width of the \textit{conv5}. $H_t$ is a recurrent internal state that is  computed jointly from  both current input ($X_t$)  and previous states encoded in $H_{t-1}$. The recurrence is computed by using a non-linear function $\varphi$, which defines exact form of the recurrent model. We exploit the long
short-term memory (LSTM) architecture \cite{Hochreiter1997} for our RNN layer. The LSTM was proposed specially to address vanishing
gradient problem, by introducing three additional multiplicative gates: the \textit{input gate}, \textit{forget gate} and \textit{output gate}. Details can be found in \cite{Hochreiter1997}. Hence the internal state in RNN hidden layer accesses the sequential context information scanned by all previous windows through the recurrent connection.  We further extend the RNN layer by using a bi-directional LSTM, which allows it to encode the recurrent context in both directions, so that the connectionist receipt field is able to cover the whole image width, e.g., 228 $\times$ width. We use a 128D hidden layer for each LSTM, resulting in a 256D RNN hidden layer, $H_t \in R^{256}$.

The internal state in $H_t$ is mapped to the following FC layer, and output layer for computing the predictions of the $t$-th proposal. Therefore, our integration with the RNN layer is elegant, resulting in an efficient model that is end-to-end trainable without additional cost. The efficiency of the RNN connection is demonstrated in Fig. \ref{fig:lstm}. Obviously, it reduces false detections considerably, and at the same time,  recovers many missed text proposals which contain very weak text information.

\subsection{Side-refinement}
The fine-scale text proposals are detected accurately and reliably by our CTPN. Text line construction is straightforward by connecting continuous text proposals whose text/non-text score is $>0.7$. Text lines are constructed as follow. First, we define a paired neighbour ($B_j$) for a proposal $B_i$ as $B_j->B_i$, when  (i) $B_j$ is the nearest horizontal distance to $B_i$, and (ii) this distance is less than 50 pixels, and (iii) their vertical overlap is $>0.7$. Second, two proposals are grouped into a pair, if $B_j->B_i$ and $B_i->B_j$. Then a text line is constructed by sequentially connecting  the pairs  having a same proposal.


 \begin{figure}[tb]
\centering
\includegraphics[height=4cm,width=12cm]{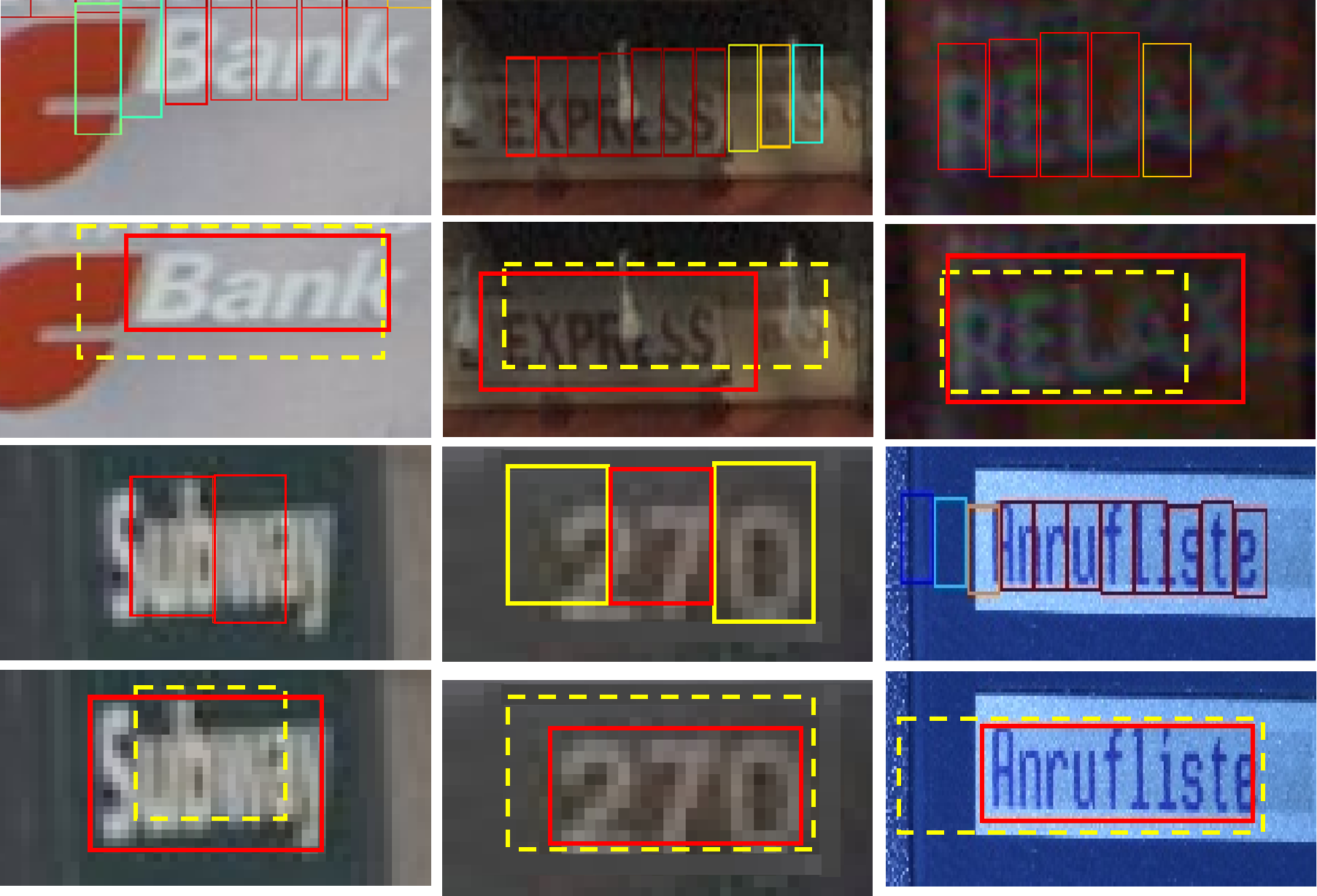}
\caption{CTPN detection with (red box) and without (yellow dashed box) the side-refinement. Color of fine-scale proposal box indicate a text/non-text score.}
\label{fig:refinement}
\end{figure}

The fine-scale detection and  RNN connection are able to predict accurate localizations in vertical direction. In horizontal direction, the image is divided into a sequence of equal 16-pixel width proposals. This  may lead to an inaccurate localization when the text proposals in both horizontal sides are not exactly covered by a ground truth text line area, or some side proposals are discarded (e.g., having a low text score), as shown in Fig. \ref{fig:refinement}. This inaccuracy may be not crucial in generic object detection, but should not be ignored in text detection, particularly for those small-scale text lines or words. To address this problem, we propose a side-refinement approach that accurately estimates the offset for each anchor/proposal in both left and right horizontal sides (referred as side-anchor or side-proposal).  Similar to the $y$-coordinate prediction, we compute relative offset as,
\begin{eqnarray}
o=(x_{side}-c_x^a)/w^a, \quad o^*=(x^*_{side}-c_x^a)/w^a 
\end{eqnarray}\label{eq:offset}
where $x_{side}$ is the predicted $x$-coordinate of the nearest horizontal side (e.g., left or right side) to current anchor. $x^*_{side}$ is the ground truth (GT) side coordinate in $x$-axis, which is pre-computed from the GT bounding box and anchor location. $c_x^a$ is the center of anchor in $x$-axis. $w^a$ is the width of anchor, which is fixed, $w^a=16$ . The side-proposals are defined as the start and end proposals when we connect a sequence of detected fine-scale text proposals into a text line. We only use the offsets of the side-proposals to refine the final text line bounding box. Several detection examples improved by side-refinement are presented in Fig. \ref{fig:refinement}. The side-refinement further improves the localization accuracy, leading to about 2\% performance improvements on the SWT and Multi-Lingual datasets. Notice that the offset for side-refinement is predicted simultaneously by our model, as shown in Fig. \ref{fig:main}. It is not computed from an additional post-processing step.

\subsection{Model Outputs and Loss Functions}
The proposed CTPN has three outputs which are jointly connected to the last FC layer, as shown in Fig. \ref{fig:main} (a). The three outputs simultaneously predict text/non-text scores ($\textbf{s}$),   vertical coordinates ($\textbf{v}=\{v_c, v_h\}$ in E.q. (\ref{eq:coordinate})) and side-refinement offset ($\textbf{o}$).
We explore $k$ anchors to predict them on each spatial location in the \textit{conv5},  resulting in $2k$, $2k$ and $k$ parameters in the output layer, respectively.

We employ multi-task learning to jointly optimize model parameters. We introduce three loss functions, $L^{cl}_s$, $L^{re}_v$ and $l^{re}_o$, which compute errors of text/non-text score, coordinate and side-refinement, respectively.  With these considerations, we follow the multi-task loss applied in  \cite{Girshick2015,Ren2015}, and minimize an overall objective function ($L$) for an image as,

\begin{eqnarray}
L(\textbf{s}_i, \textbf{v}_j, \textbf{o}_k) =\frac1{N_{s}}\sum_iL^{cl}_{s}(\textbf{s}_i, \textbf{s}_i^*)
+\frac{\lambda_1}{N_v}\sum_j L^{re}_v(\textbf{v}_j, \textbf{v}_j^*)
+\frac{\lambda_2}{N_o}\sum_k L^{re}_o(\textbf{o}_k, \textbf{o}_k^*)
\end{eqnarray}
where each anchor is a training sample, and $i$ is the index of an anchor in a mini-batch. $\textbf{s}_i$ is the predicted probability of anchor $i$ being a true text. 
$\textbf{s}_i^*=\{0,1\}$ is the ground truth. 
$j$ is the index of an anchor in the set of valid anchors for $y$-coordinates regression, which are defined as follow.
A valid anchor is a defined positive anchor ($\textbf{s}_j^*=1$, described below), or has an Intersection-over-Union (IoU) $>0.5$ overlap  with a ground truth text proposal.
$\textbf{v}_j$ and $\textbf{v}_j^*$ are the prediction and ground truth $y$-coordinates associated with the $j$-{th} anchor.
$k$ is the index of a side-anchor, which is defined as a set of anchors within a horizontal distance (e.g.,  32-pixel) to the left or right side of a ground truth text line bounding box.
$\textbf{o}_k$ and $\textbf{o}_k^*$ are the predicted and ground truth offsets in $x$-axis associated to the $k$-{th} anchor.
$L^{cl}_s$ is the classification loss which we use Softmax loss to distinguish text and non-text. $L^{re}_v$ and $L^{re}_o$ are the regression loss. 
We follow previous work by using the smooth $L_1$ function to compute them \cite{Girshick2015,Ren2015}.
$\lambda_1$ and $\lambda_2$ are loss weights to balance different tasks, which are empirically set to 1.0 and 2.0.  $N_{s}$ $N_{v}$ and $N_{o}$ are normalization parameters, denoting the total number of anchors used by $L^{cl}_s$, $L^{re}_v$ and $L^{re}_o$, respectively.


\subsection{Training and Implementation Details}
The CTPN can be trained end-to-end by using the standard back-propagation and stochastic gradient descent (SGD). Similar to RPN \cite{Ren2015}, training samples are the anchors, whose locations can be pre computed in input image, so that the training labels of each anchor can be computed from corresponding GT box.

\textbf{Training labels}.  For text/non-text classification,  a binary label is assigned to each positive (text) or negative (non-text) anchor. It is defined by computing the IoU overlap with the GT bounding box (divided by anchor location). A positive anchor is defined as : (i) an anchor that has an $>0.7$ IoU overlap with any GT box; \textit{or} (ii) the anchor with the highest IoU overlap with a GT box. \textit{By the condition (ii), even a very small text pattern can assign a positive anchor. This is crucial to detect small-scale text patterns, which is one of key advantages of the CTPN. } This is different from generic object detection where the impact of condition (ii) may be not significant. The negative anchors are defined as $<0.5$ IoU overlap with all GT boxes. The training labels for the $y$-coordinate regression ($\textbf{v}^*$) and offset regression  ($\textbf{o}^*$) are computed as E.q. (\ref{eq:coordinate_gt}) and (4) respectively.

\textbf{Training data.} In the training process, each mini-batch samples are collected randomly from a single image. The number of anchors for each mini-batch is fixed to  $N_s=128$, with 1:1 ratio for positive and negative samples. A mini-patch is pad with negative samples if the number of positive ones is fewer than 64. Our model was trained on 3,000 natural images, including 229 images from the ICDAR 2013 training set. We collected the other images ourselves and manually labelled them with text line bounding boxes. All self-collected training images are not overlapped with any test image in all benchmarks.  The input image is resized by setting its short side to 600 for training, while keeping its original aspect ratio.

\textbf{Implementation Details.} We follow the standard practice,  and explore the  very deep VGG16 model \cite{Simonyan2015} pre-trained on the ImageNet data \cite{Russakovsky2015}. We initialize the new layers (e.g., the RNN and output layers) by using random weights with Gaussian distribution of 0 mean and 0.01 standard deviation.  The model was trained end-to-end by fixing the parameters in the first two convolutional layers. We used 0.9 momentum and 0.0005 weight decay. The learning rate was set to 0.001 in the first 16K iterations, followed by another 4K iterations with 0.0001 learning rate. Our model was implemented in Caffe framework \cite{Jia2014}.

\section{Experimental Results and Discussions}
We evaluate the  CTPN on five text detection benchmarks, namely the ICDAR 2011 \cite{Minetto2010}, ICDAR 2013 \cite{Karatzas2013}, ICDAR 2015 \cite{Karatzas2015}, SWT \cite{Epshtein2010}, and Multilingual dataset \cite{Pan2011}. In our experiments, we first verify the efficiency of each proposed component individually, e.g., the fine-scale text proposal detection or  in-network recurrent connection. The ICDAR 2013 is used for this component evaluation.

\subsection{Benchmarks and Evaluation Metric} 
\textit{The ICDAR 2011} dataset \cite{Minetto2010} consists of 229 training images and 255 testing ones, where the images are labelled in word level. \textit{The ICDAR 2013 }\cite{Karatzas2013} is similar as the ICDAR 2011, and has in total 462 images, including 229 images and 233 images for training and testing, respectively.  \textit{The ICDAR 2015} (Incidental Scene Text - Challenge 4) \cite{Karatzas2015} includes 1,500 images which were collected by using the Google Glass. The training set has 1,000 images, and the remained 500 images are used for test. This dataset is more challenging than previous ones by including arbitrary orientation, very small-scale and low resolution text. \textit{The Multilingual} scene text dataset is collected by \cite{Pan2011}. It contains 248 images for training and 239 for testing. The images include multi-languages text, and the ground truth is labelled in text line level. Epshtein \textit{et al.} \cite{Epshtein2010} introduced \textit{the SWT} dataset containing 307 images which include many extremely small-scale text.

We follow previous work by using standard evaluation protocols which are provided by the dataset creators or competition organizers. For the ICDAR 2011 we use the standard protocol proposed by \cite{Wolf2006},  the evaluation on the ICDAR 2013 follows the standard in \cite{Karatzas2013}. For the ICDAR 2015, we used the online evaluation system provided by the organizers as in \cite{Karatzas2015}. The evaluations on the SWT and Multilingual datasets follow the  protocols defined in \cite{Epshtein2010} and  \cite{Pan2011} respectively.  



\subsection{Fine-Scale Text Proposal Network with Faster R-CNN}

We first discuss our fine-scale detection strategy against the RPN and Faster R-CNN system \cite{Ren2015}.  As can be found in Table \ref{tab:component_swt} (left), the individual RPN is difficult to perform accurate text localization, by generating a large amount of false detections (low precision). By refining the RPN proposals with a Fast R-CNN detection model \cite{Girshick2015}, the Faster R-CNN system improves  localization accuracy considerably, with a F-measure of 0.75.  One observation is that the Faster R-CNN also increases the recall of original RPN. This may benefit from joint bounding box regression mechanism of the Fast R-CNN, which improves the accuracy of a predicted bounding box. The RPN proposals may roughly localize a major part of a text line or word, but they are not accurate enough by the ICDAR 2013  standard. Obviously, the proposed fine-scale text proposal network (FTPN) improves the Faster R-CNN remarkably in both precision and recall, suggesting that the FTPN is more accurate and reliable,  by predicting a sequence of fine-scale text proposals rather than a whole  text line. 

\subsection{Recurrent Connectionist Text Proposals}
We discuss  impact of recurrent connection on our CTPN. As shown in Fig. \ref{fig:lstm}, the context information is greatly helpful to reduce false detections, such as text-like outliers. It is of great importance for recovering highly ambiguous text (e.g., extremely small-scale ones), which is one of main advantages of our CTPN, as demonstrated in Fig. \ref{fig:samll}. These appealing properties result in a significant performance boost. As shown in Table \ref{tab:component_swt} (left), with our recurrent connection, the CTPN improves the FTPN substantially from a F-measure of 0.80 to 0.88.


\textbf{Running time.} The implementation time of our CTPN (for whole detection processing) is about $0.14s$ per image with a fixed short side of 600, by using a single GPU. The CTPN without the RNN connection takes about 0.13s/image GPU time. Therefore, the proposed  in-network recurrent mechanism increase model computation  marginally, with considerable performance gain obtained. \\

\begin{minipage}[b]{0.9\linewidth}
\centering
\captionof{table}{Component evaluation on the ICDAR 2013, and State-of-the-art results on the SWT and MULTILINGUAL.}\label{tab:component_swt} 
\begin{adjustbox}{max width=\textwidth}
\begin{tabular}{l|*{3}{c|}|l|*{3}{c}||l*{3}{|c}}

\hline
\multicolumn{4}{c||}{Components on ICDAR 2013}
&\multicolumn{4}{|c||}{SWT}
&\multicolumn{4}{|c}{MULTILINGUAL}
\\
\hline
\multirow{1}{*}{Method} 
&\multicolumn{1}{|c}{P}
&\multicolumn{1}{|c}{R}
&\multicolumn{1}{|c||}{F} 
&\multirow{1}{*}{Method} 
&\multicolumn{1}{|c}{P}
&\multicolumn{1}{|c}{R}
&\multicolumn{1}{|c||}{F} 
&\multirow{1}{*}{Method} 
&\multicolumn{1}{|c}{P}
&\multicolumn{1}{|c}{R}
&\multicolumn{1}{|c}{F}

\\
\hline
\hline
RPN & 0.17 & 0.63 & 0.27 & Epshtein \cite{Epshtein2010}  &  0.54   & 0.42     &  0.47  & Pan \cite{Pan2011}  & 0.65 & 0.66 & 0.66 \\
Faster R-CNN & 0.79 & 0.71 & 0.75 & Mao  \cite{Mao2013}  & 0.58 & 0.41 & 0.48 & Yin \cite{Yin2014}  &  0.83 & 0.68 & 0.75\\
FTPN (no RNN) &0.83 & 0.78 &0.80 &  Zhang \cite{Zhang2015}  & \textbf{0.68}       &0.53   &0.60 & Tian \cite{Tian2015} & \textbf{0.85} & 0.78 & 0.81  \\
\hline
\hline
CTPN & \textbf{0.93} & \textbf{0.83} & \textbf{0.88} & CTPN & \textbf{0.68} & \textbf{0.65} & \textbf{0.66} & CTPN  &  0.84 & \textbf{0.80} &  \textbf{0.82} \\
\hline

\end{tabular}
\end{adjustbox}
\bigskip
\end{minipage}

\subsection{Comparisons with state-of-the-art results}
\begin{figure}[tb]
\centering
\includegraphics[height=5cm,width=12cm]{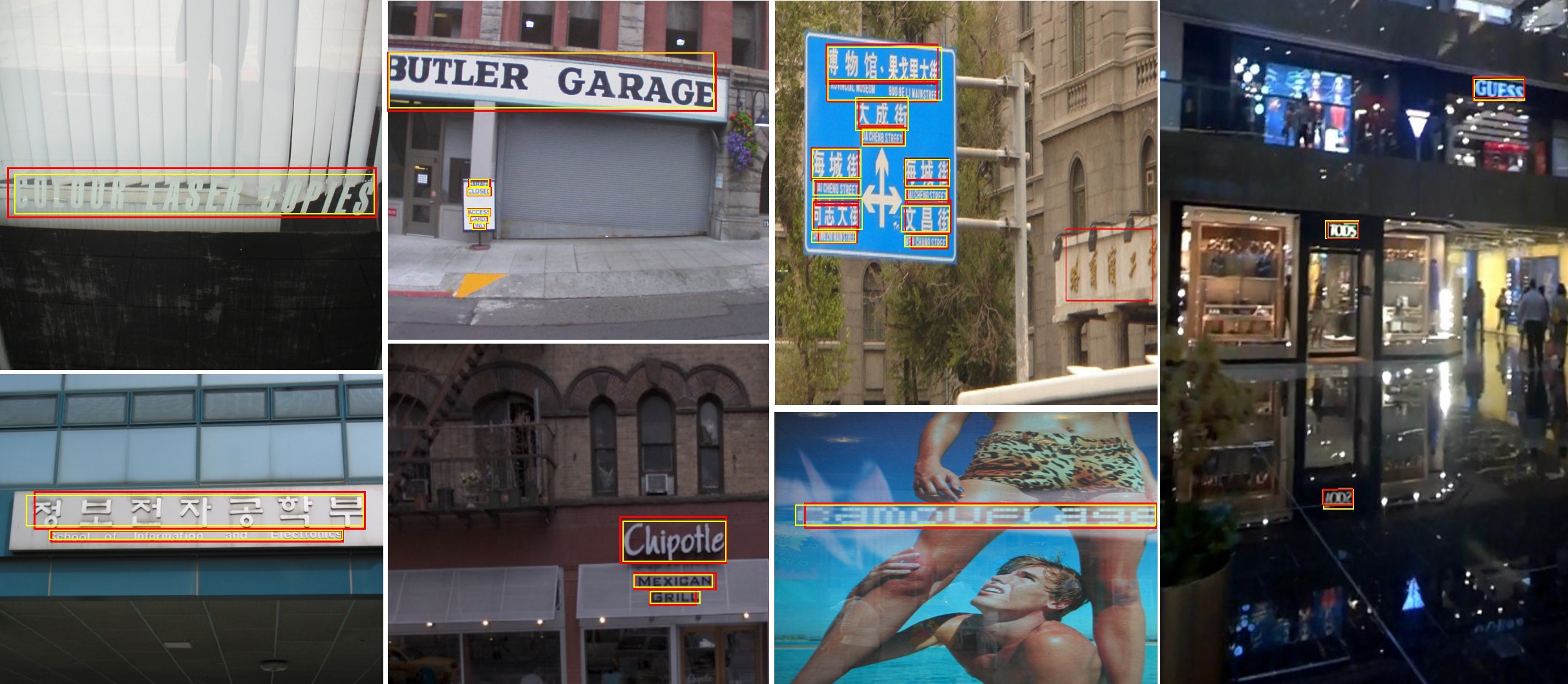}
\caption{CTPN detection results several challenging images, including multi-scale and multi-language text lines. Yellow boxes are the ground truth. }
\label{fig:final}
\end{figure}

 Our detection results on several challenging images are presented in Fig. \ref{fig:final}.  As can be found, the CTPN works perfectly on these challenging cases, some of which are difficult for many previous methods.  It is able to handle multi-scale and multi-language efficiently (e.g., Chinese and Korean). \\
 
\begin{minipage}[b]{0.9\linewidth}
\centering
\captionof{table}{State-of-the-art results on the ICDAR 2011, 2013 and 2015.}\label{tab:icdar} 
\begin{adjustbox}{max width=\textwidth}
\begin{tabular}{l|*{3}{c|}|l|*{3}{c|}{c}||l*{3}{|c}}
\hline
\multicolumn{4}{c||}{ICDAR 2011}
&\multicolumn{5}{|c||}{ICDAR 2013}
&\multicolumn{4}{|c}{ICDAR 2015}
\\
\hline
\multirow{1}{*}{Method}
&\multicolumn{1}{|c}{P}
&\multicolumn{1}{|c}{R}
&\multicolumn{1}{|c||}{F} 
&\multirow{1}{*}{Method}
&\multicolumn{1}{|c}{P}
&\multicolumn{1}{|c}{R}
&\multicolumn{1}{|c}{F}
&\multicolumn{1}{|c||}{T($s$)}
&\multirow{1}{*}{Method}
&\multicolumn{1}{|c}{P}
&\multicolumn{1}{|c}{R}
&\multicolumn{1}{|c}{F} 

\\
\hline
\hline
Huang \cite{Huang2013}  & 0.82 & 0.75 & 0.73  & Yin \cite{Yin2014}  &0.88  & 0.66 & 0.76 &0.43 & CNN Pro. & 0.35 & 0.34 & 0.35\\
Yao \cite{Yao2014}  & 0.82 & 0.66 & 0.73  &Neumann \cite{Neumann2015b}  & 0.82& 0.72 & 0.77 &0.40 &Deep2Text &0.50 & 0.32& 0.39\\
Huang \cite{Huang2014}  & 0.88 & 0.71 & 0.78  & Neumann  \cite{Neumann2015}  & 0.82& 0.71 & 0.76 &0.40 & HUST & 0.44 & 0.38 & 0.41 \\
Yin  \cite{Yin2014}  & 0.86 & 0.68 & 0.76  &  FASText \cite{Busta2015}  & 0.84 & 0.69 & 0.77&0.15 & AJOU & 0.47 & 0.47 & 0.47 \\
Zhang  \cite{Zhang2015}   &0.84       &0.76    &0.80 & Zhang  \cite{Zhang2015}   & 0.88 & 0.74 & 0.80 &60.0 & NJU-Text & 0.70 & 0.36 & 0.47 \\
TextFlow \cite{Tian2015}  & 0.86 & 0.76 & 0.81 & TextFlow \cite{Tian2015}  & 0.85 & 0.76 & 0.80 &0.94 & StradVision1& 0.53 & 0.46 & 0.50   \\
Text-CNN \cite{He2016}  & 0.91 & 0.74 & 0.82 & Text-CNN \cite{He2016}  & 0.93 & 0.73 & 0.82 &4.6 & StradVision2 & 0.77 & 0.37 & 0.50  \\
Gupta \cite{Gupta2016} & \textbf{0.92} &0.75&0.82 & Gupta \cite{Gupta2016} & 0.92 & 0.76 & 0.83 & \textbf{0.07}& Zhang \cite{Zhang2016} & 0.71 & 0.43 & 0.54 \\
\hline
\hline
CTPN  & 0.89 & \textbf{0.79} & \textbf{0.84} & CTPN  & \textbf{0.93} & \textbf{0.83} &  \textbf{0.88} & 0.14 $^*$ & CTPN & \textbf{0.74} & \textbf{0.52} &\textbf{0.61} \\
\hline

\end{tabular}
\end{adjustbox}
\bigskip
\end{minipage}

\begin{figure}[tb]
\centering
\includegraphics[height=4.8cm,width=12cm]{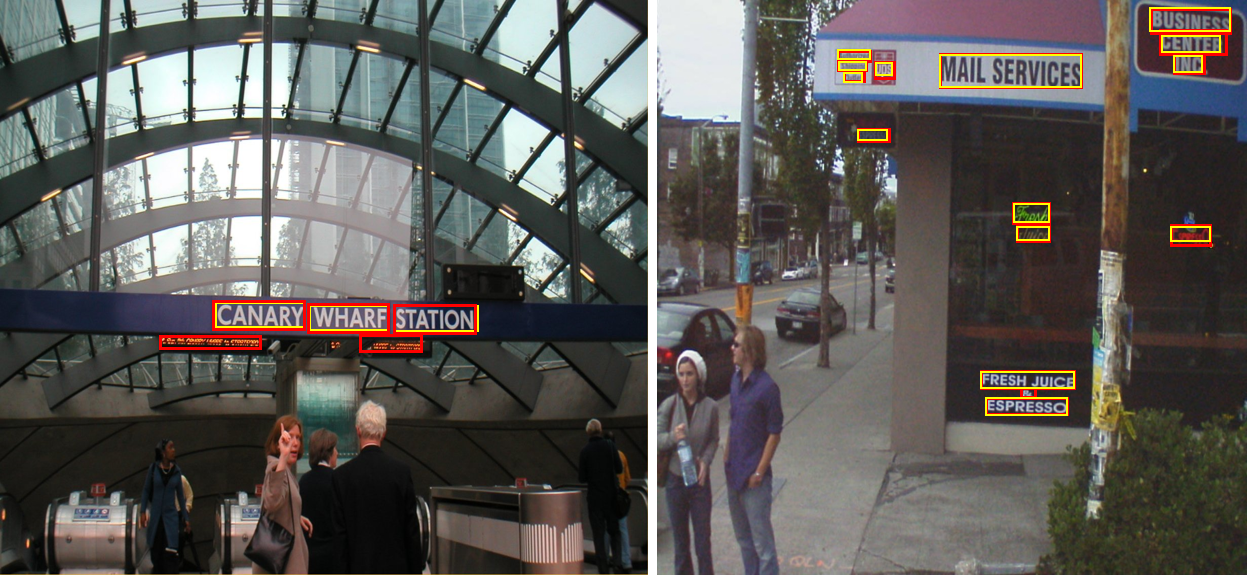}
\caption{CTPN detection results on extremely small-scale cases (in red boxes), where some ground truth boxes are missed. Yellow boxes are the ground truth. }
\label{fig:samll}
\end{figure}

The full evaluation was conducted on five benchmarks. Image resolution  is varied significantly in different datasets. We set short side of images to 2000 for the SWT and ICDAR 2015, and 600 for the other three. We compare our performance against recently published results in \cite{Busta2015,Tian2015,Zhang2015}. As shown in Table \ref{tab:component_swt} and \ref{tab:icdar}, our CTPN achieves the best performance on all five datasets. On the SWT, our improvements  are  significant on both recall and F-measure, with  marginal gain on precision. Our detector performs favourably against the TextFlow on the Multilingual, suggesting that our method generalize well to various languages. On the  ICDAR 2013, it outperforms recent TextFlow \cite{Tian2015} and FASText \cite{Busta2015} remarkably by improving the F-measure from 0.80 to 0.88. The gains are considerable in both precision and recall, with more than $+5\%$ and $+7\%$ improvements, respectively. 
 In addition, we further compare our method against \cite{Gupta2016,He2016,Zhang2016}, which were published after our initial submission. It consistently obtains substantial  improvements on F-measure and recall. 
 This may due to  strong capability of  CTPN for detecting extremely challenging text, e.g., very small-scale ones,  some of which are even difficult for human.  As shown in Fig. \ref{fig:samll}, those challenging ones are detected correctly by our detector, but some of them  are even missed by the GT labelling, which may reduce our precision in evaluation.




 We further investigate running time of various  methods, as compared in Table \ref{tab:icdar}. FASText \cite{Busta2015} achieves $0.15s$/image CPU time. Our method is slightly faster than it by obtaining $0.14s$/image, but in GPU time. Though it is not fair to  compare them directly, the GPU computation has become mainstream with recent great success of deep learning approaches on object detection \cite{Ren2015,Girshick2015,Girshick2014}. Regardless of running time, our method outperforms the FASText substantially with 11\% improvement on F-measure. Our time can be reduced by using a smaller image scale. By using the scale of 450, it is reduced to $0.09s$/image, while obtaining P/R/F of 0.92/0.77/0.84 on the ICDAR 2013, which are compared competitively against Gupta \textit{et al.}'s approach \cite{Gupta2016} using $0.07s$/image with GPU.



\section{Conclusions}

We have presented a Connectionist Text Proposal Network (CTPN) - an efficient text detector that is end-to-end trainable. The CTPN detects a text line in a sequence of fine-scale text proposals directly in convolutional maps. We develop vertical anchor mechanism that jointly predicts precise location and text/non-text score for each proposal, which is the key to realize \textit{accurate} localization of text. 
We propose an in-network RNN layer that connects sequential text proposals elegantly, allowing it to explore meaningful context information. These key technical developments result in a powerful ability to detect highly challenging text, with less false detections. The CTPN is efficient by achieving new state-of-the-art performance on five benchmarks, with $0.14s$/ image running time.


\clearpage

\bibliographystyle{splncs03}
\bibliography{egbib2}
\end{document}